%% file: main_ICLP.tex
\documentclass[submission,copyright,creativecommons]{eptcs}
\providecommand{\event}{ICLP 2025} 

\usepackage{iftex}

\ifpdf
  \usepackage{underscore}         
  \usepackage[T1]{fontenc}        
\else
  \usepackage{breakurl}           
\fi

\usepackage{listings}
\usepackage{graphicx}
\usepackage{booktabs}
\usepackage{amsmath}
\usepackage{algorithm}
\usepackage{algorithmic}
\usepackage{url}
\usepackage{hyperref}
\usepackage{appendix}
\usepackage{xcolor}

\title{Interpretable Hybrid Machine Learning Models Using
FOLD-R++ and Answer Set Programming}
\def\titlerunning{Hybrid ML using FOLD-R++ and ASP}
\def\authorrunning{S. Wielinga and J. Heyninck}
\author{Sanne Wielinga 
\institute{Open Universiteit, the Netherlands}
\email{sannwielinga@gmail.com}
\and
Jesse Heyninck
\institute{Open Universiteit \\ Heerlen, Limburg, the Netherlands}
\institute{University of Cape Town and CAIR\\
Cape Town, South Africa}
\email{jesse.heyninck@ou.nl}
}

\begin{document}

\maketitle

\input{content}

\subsection*{Acknowledgements}

This work is partially supported by the project {LogicLM}: Combining Logic Programs with Language Model with the number NGF.1609.241.010 of the research programme NGF AiNed XS Europa 2024-1 which is (partly) financed by the Dutch Research Council (NWO).
\bibliographystyle{eptcsalpha}
\bibliography{references}

\begin{appendices}
\input{appendix}
\end{appendices}

\end{document}

%% file: content.tex
\lstset{
    language=Prolog,     
    basicstyle=\small\ttfamily,
    columns=fullflexible,
    breaklines=true,      
    frame=single,
    captionpos=b,       
}

\begin{abstract}
    Machine learning (ML) techniques play a pivotal role in high-stakes domains such as healthcare, where accurate predictions can greatly enhance decision-making. However, most high-performing methods such as neural networks and ensemble methods are often opaque, limiting trust and broader adoption. In parallel, symbolic methods like Answer Set Programming (ASP) offer the possibility of interpretable logical rules but do not always match the predictive power of ML models. This paper proposes a hybrid approach that integrates ASP-derived rules from the FOLD-R++ algorithm with black-box ML classifiers to selectively correct uncertain predictions and provide human-readable explanations. Experiments on five medical datasets reveal statistically significant performance gains in accuracy and F1 score. This study underscores the potential of combining symbolic reasoning with conventional ML to achieve high interpretability without sacrificing accuracy. 
\end{abstract}


\section{Introduction}

Machine learning has become an essential tool for predictive analytics and decision-making in various domains, including healthcare \cite{tonekaboni2019clinicians}. Advanced ML models, particularly black-box models such as neural networks and ensemble methods, have shown impressive predictive performance. However, many of these high-performing models offer little insight into \emph{why} a specific prediction is made. This lack of transparency slows down the adoption of models in applications where understanding the reasoning behind a prediction is as important as the prediction itself \cite{arrieta2020explainable}. In fields like medicine, decisions informed by ML models can have serious implications on patient outcomes. The ability to explain and justify predictions is therefore important for trust, accountability, and informed decision-making \cite{doshi2017towards}.

Meanwhile, symbolic methods such as Answer Set Programming (in short, ASP), allow to represent knowledge in transparent, human-understandable logical rules \cite{erdem2016applications}. The FOLD-R++ algorithm uses ASP to learn default rules \cite{wang2022fold}. These rules can complement black-box models by offering explanations for their predictions and therefore improving interpretability.

However, integrating ASP-derived rules with black-box ML models without changing their internal mechanisms remains challenging. Hybrid neuro-symbolic methods have emerged as a promising solution, though most of them emphasize neural architectures and require specialized integration \cite{garcez2019neural,manhaeve2018deepproblog,yang2023neurasp}. In contrast, relatively few works integrate symbolic rules with other types of black-box models, such as support vector machines or ensemble methods. 

This paper proposes a hybrid approach that addresses this gap by integrating FOLD-R++, an algorithm that learns ASP rules with exceptions \cite{wang2022fold}, with various black-box ML models. By using rules that capture domain knowledge, the hybrid system can (1) correct ML model errors when the model is uncertain and (2) provide human-readable explanations.

The following research questions guide this study:
\begin{enumerate}
\item[] \textbf{RQ1}: Does integrating interpretable ASP rules derived from FOLD-R++ improve the predictive performance of black-box ML models across various medical datasets?
\item[] \textbf{RQ2}: How does the hybrid model of ML and ASP improve the interpretability of predictions?
\end{enumerate}

The remainder of this paper is organized as follows: Section~\ref{sec:related-work} provides an overview of interpretability research and other related approaches. Section~\ref{sec:background} covers the background on ASP syntax and semantics, the FOLD-R++ algorithm, and a brief discussion of traditional ML methods. Section~\ref{sec:methods} outlines the methodology, including data preparation, model training, and hybrid model implementation. Section~\ref{sec:results} presents the empirical findings. Section~\ref{sec:conclusion} concludes with limitations and paths for future research.


\section{Related Work}
\label{sec:related-work}

\subsection{Interpretability and Related Approaches}

The need for interpretable ML has surged amid concerns for trust, transparency, and accountability \cite{guidotti2018survey}. Traditional interpretable models, such as decision trees and linear models, provide clear decision paths but may miss complex interactions in large datasets. In contrast, advanced ML techniques, such as deep neural networks and ensemble methods, achieve superior performance but function as "black-box" models with opaque decision-making processes.. This opacity is problematic in areas such as healthcare, where interpretability is essential for ethical considerations, regulatory compliance, and establishing trust among clinicians and patients \cite{tonekaboni2019clinicians}. Clinicians must understand how a model arrived at a specific diagnosis to trust and effectively use it in decision-making \cite{tonekaboni2019clinicians}. Lack of transparency can lead to mistrust or rejection of ML systems \cite{arrieta2020explainable}. Moreover, regulatory bodies increasingly require explanations for automated decisions, especially when they impact patient care \cite{palaniappan2024global}. 

Researchers have proposed various strategies to address these challenges. These include developing inherently interpretable models such as decision trees and rule-based systems \cite{rudin2019stop}, and using post-hoc techniques like 
SHAP \cite{lundberg2017unified} to explain predictions from black-box models. These methods attempt to approximate the behavior of complex models locally with simpler models. However, a trade-off exists between interpretability and performance: simple models may miss complex patterns, while complex models remain opaque. 
Our proposed hybrid method aims to navigate this trade-off by integrating interpretable rules with high-performing black box models.

\subsection{Hybrid Models}

A key line of related work focuses on combining symbolic reasoning and machine learning, which focuses on integrating symbolic reasoning with neural networks, creating neuro-symbolic systems \cite{garcez2019neural}. For instance, DeepProbLog integrates probabilistic logic programming with deep learning \cite{manhaeve2018deepproblog}. NeurASP is a framework that combines neural networks with ASP \cite{yang2023neurasp}.

These approaches often require modifications to the learning algorithms or network architectures, which can be complex and computationally intensive. Moreover, they primarily focus on neural networks and do not address integration with other types of black-box ML models, such as random forests or support vector machines.
Thus, a gap remains on the \emph{model-agnostic}  integration of symbolic reasoning.

This study addresses this gap by integrating interpretable ASP rules from FOLD-R++ with a variety of black-box ML classifiers. The hybrid approach does not require changing the ML models, thus preserving their performance. The ASP component adds a layer of interpretability by offering human-understandable explanations for the predictions.


\section{Background}
\label{sec:background}

\subsection{Symbolic Logic and Answer Set Programming}

Symbolic logic offers a framework for interpretable models by encoding knowledge in structured, human-readable formats. Answer Set Programming combines logic programming with non-monotonic reasoning to represent complex relationships and constraints \cite{erdem2016applications}. ASP programs consist of rules, which in turn are composed of literals (positive or negated atomic formulas) and constraints that define which combinations of literals cannot all be true simultaneously (integrity constraints). A key aspect of ASP is its use of stable model or answer set semantics to determine which sets of ground literals are considered stable solutions to a program. The non-monotonic nature of ASP makes it ideal for representing default rules and exceptions.
An answer set program \text{II} consists of rules of the form:

\texttt{a\textsubscript{0} :- a\textsubscript{1}, ..., a\textsubscript{m}, not a\textsubscript{m+1}, ..., not a\textsubscript{n}.}

where each \texttt{a\textsubscript{i}} is an atom, and  \texttt{not} represents negation as failure. The head of the rule (\texttt{a\textsubscript{0}}) is derived if all positive literals in the body (\texttt{a\textsubscript{1}, ...,a\textsubscript{m}}) are true and all negated literals (\texttt{not a\textsubscript{m+1},..., not a\textsubscript{n}}) are not proven true. An answer set is a minimal set of literals satisfying all program rules without cyclic support. Multiple answer sets represent different possible solutions to the encoded problem. Solvers like clingo \cite{gebser2019multi} can automatically calculate these answer sets.

ASP is particularly valuable for tasks requiring interpretability. Applications of ASP can be found in many areas, including planning, scheduling, and bio-informatics \cite{erdem2016applications}. In healthcare specifically, ASP has been used to model clinical guidelines \cite{spiotta2017temporal} and solve scheduling problems \cite{erdem2016applications}.

\subsection{FOLD-R++ Algorithm}

The FOLD-R++ algorithm \cite{wang2022fold} extends the First-Order Logical Decision tree (FOLD) algorithm to learn default rules with exceptions from relational data, producing output that is both human-readable and compatible with ASP reasoning.
FOLD-R++ generates default rules for general patterns while capturing exceptions for special cases or anomalies. It operates by recursively partitioning the data to construct a decision tree (similar to algorithms like ID3 or C4.5), but transforms the decision tree into an ASP.

The algorithm starts with the complete dataset and evaluates all possible literals (attribute-value pairs or relational literals) that could split the data. At each node, it selects the literal that best separates positive from negative examples according to a heuristic, incorporating this literal into the rule condition. When a rule cannot perfectly classify the data, the algorithm identifies exceptions to the rule and induces them as sub-rules through the same recursive process. Mechanisms for pruning unnecessary rules or exceptions are included to prevent overfitting. The final ruleset is translated into an ASP program.

FOLD-R++-induced rules have the following form:
$$\verb|label(X, Class) :- Conditions(X), not Exceptions(X).|$$
 where \texttt{Conditions(X)} represent the conditions under which an instance \texttt{X} belongs to a particular class \verb|Class|, and \texttt{Exceptions(X)} are exceptions to this rule.

We chose FOLD-R++ for its proven effectiveness and compatibility with the experimental setup.
Although newer algorithms in the same family, like FOLD-SE and FOLD-RM, offer advanced features, they were not selected due to accessibility constraints and alignment with the research focus. 

\subsection{Traditional ML Methods}
\label{section:traditional:ml}
In addition to FOLD-R++, this paper considers four widely used machine learning classifiers:

\begin{itemize}
    \item \textbf{Random Forest (RF):} An ensemble method building multiple decision trees using bootstrap aggregation and random feature selection, reducing variance and improving generalization.
    \item \textbf{Support Vector Machine (SVM):} A model that finds the optimal hyperplane that best separates classes in a high-dimensional space.
    \item \textbf{K-Nearest Neighbors (KNN):} A non-parametric method that classifies instances based on the majority class among the k-nearest neighbors in the feature space.
    \item \textbf{Neural Network (MLPClassifier):} A multi-layer perceptron that learns complex non-linear relationships through backpropagation.
\end{itemize}

These models represent a spectrum of approaches in modern machine learning, from relatively transparent (KNN) to highly opaque (neural networks). Each has strengths and weaknesses in terms of predictive performance, generalization ability, and interpretability. Of course, this is just a selection of ML-methods, and our framework allows for the integration of other ML-classifiers as well.


\section{Methods}
\label{sec:methods}

\begin{figure}[tb]
    \centering
    \includegraphics[scale=0.5]{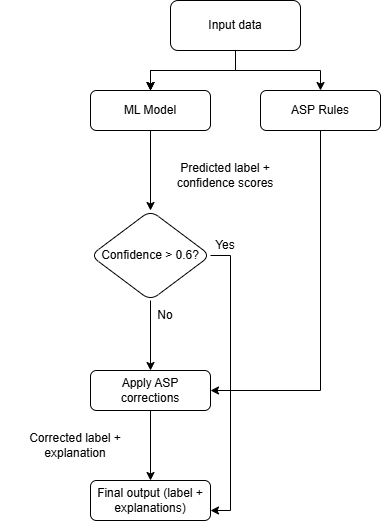}
    \vspace{10pt}
    \caption{Conceptual diagram of the hybrid model architecture. Input data is processed by both the black-box ML model and the ASP rule-based component. The confidence score determines whether the ML prediction is accepted directly or corrected by the ASP rules.}
    \label{fig:diagram}
\end{figure}

This section outlines the methodology used to develop and evaluate the hybrid models that combine black-box ML models with interpretable rules generated by FOLD-R++ using ASP. The general pipeline consists of data preparation, ML model training, rule induction via FOLD-R++, and final integration of predictions using a confidence threshold. Figure~\ref{fig:diagram} provides an overview of this pipeline. Relevant data and code can be found {online}: \url{https://github.com/sannewielinga/foldrpp-asp-hybrid}.

\subsection{Data Preparation}

Five medical datasets were selected from the UCI Machine Learning Repository \cite{ucirepository}:

\begin{itemize}
\item \textbf{Heart Disease:} Contains 303 instances with 14 commonly used attributes. It is used to predict the presence of heart disease in patients based on medical measurements.
\item \textbf{Autism Screening Adult:} Includes 704 instances with 21 attributes, used to predict the likelihood of autism spectrum disorder based on screening test scores and demographic information.
\item \textbf{Breast Cancer Wisconsin:} Consists of 569 instances with 30 numerical features computed from digitized images. The goal is to classify tumors as malignant or benign.
\item \textbf{Ecoli:} Contains 336 instances with 8 attributes, used to classify protein localization sites.
\item \textbf{Chronic Kidney Disease:} Contains 400 instances with 24 attributes, used to predict the presence of chronic kidney disease in patients based on clinical and laboratory findings.
\end{itemize}

Each dataset underwent preprocessing to handle missing values, encode categorical variables, and scale numerical features where necessary. Stratified sampling was used to split the datasets into training and testing sets to maintain class distribution. A different random seed was used for each experiment.

\subsection{FOLD-R++}

FOLD-R++ \cite{wang2022fold} was used to induce ASP rules from the training data. The original FOLD-R++ code was used and slightly refactored to fit the experimental setup. This included adding new functions to transform data into ASP-compatible formats and interface with Clingo \cite{gebser2019multi}. Additionally, a wrapper was created to handle the training of FOLD-R++, conversion of induced rules to ASP syntax, and prediction using Clingo.
The training data was transformed into a format compatible with FOLD-R++, where each instance was represented as a set of logical facts with attributes and their values forming predicates. For example, an attribute-value pair such as \texttt{age = 45} was converted into a predicate \texttt{age(X,45)}.

The induced rules were converted into ASP syntax compatible with the Clingo solver. Numeric values were scaled by a factor 10 to handle the limitations of Clingo with floating-point numbers. The scaling was applied consistently to keep relationships between variables valid. Clingo was then used to apply the logical rules to the test data, reasoning over the facts representing each instance and the induced rules to infer the class labels.

\subsection{ML Models}

Four black-box ML models were trained for each dataset: Random Forest, SVM, KNN, and MLPClassifier (see Section \ref{section:traditional:ml}).
Default hyperparameters were used for all models unless adjustments were necessary for convergence or performance. Random seeds were set for model initialization and data splitting to ensure reproducibility across experiments. For models that provide probability estimates (e.g., RF, SVM with \texttt{probability=True}, MLP), the \texttt{predict\_proba} method was used to obtain confidence scores for each class. For KNN, probabilities were derived from the proportion of neighbors belonging to each class.

\subsection{Hybrid Model Implementation}

The hybrid model integrates predictions from the ML models with the interpretable rules generated by FOLD-R++ using ASP. First, predictions and confidence scores from the ML model on the test set were obtained. Clingo was then used to apply the induced rules to the test instances and generate predictions.

A fixed 0.6 confidence threshold, empirically selected from preliminary evaluations, was used to determine when to rely on the prediction of the ML model versus the ASP rules. If the confidence score of the ML model for a prediction was above 0.6, the ML prediction was used; otherwise, the prediction of the ASP rule was used. This approach ensures that the ASP rules intervene specifically when the ML model is uncertain. During development, both dynamic and fixed confidence thresholds were experimented with, but the dynamic threshold led to overfitting on the training data. In practice, this threshold can be tuned based on domain needs.

For each prediction corrected by the ASP rules, explanations were generated from the induced rules. These explanations highlight which logical conditions were met and which exceptions were not triggered. Figure~\ref{fig:diagram} provides a schematic representation of the hybrid model architecture. 

\subsection{Experimental Setup}

Ten experiments were conducted for each combination of dataset and model. Each experiment used stratified sampling for an 80-20 train-test split with varying random seeds to alter splits and model initialization. These represented repeated random subsamples rather than traditional k-fold cross-validation, testing result robustness through split variability. 
For each experiment, accuracy, precision, recall, and F1 score (the harmonic mean of precision and recall) were computed and averaged across the ten experiments (with standard deviations calculated).

Paired t-tests were conducted to determine whether the improvements observed with the hybrid model were statistically significant compared to the standalone ML models. A p-value threshold of 0.05 was used to determine statistical significance. If the ML model and hybrid model had the same accuracy in all experiments, the differences were zero, and the paired t-test could not be performed.

The importance of each ASP rule was determined based on how frequently it corrected errors made by the ML models. Due to variability in the induced ASP programs across different experiments - likely caused by randomness in data splitting - only the ASP programs and important rules from model-dataset combinations that showed statistically significant improvements were saved. Explanations were generated for instances where the hybrid model corrected the prediction of the ML model, using proof trees from the FOLD-R++ algorithm.
For example, in one instance from the Heart Disease dataset (\texttt{patient5}), the SVM model predicted a positive label (indicating heart disease), which the hybrid model corrected this to a negative prediction. The FOLD-R++ explanation is summarized as follows:\\

 \noindent $\bullet$ \textbf{Extracted Features:} \texttt{\{'major\_vessels': '2', 'chest\_pain': '4', 'age': 60.0, 'slope': '2', 'blood\_pressure': 150.0, 'oldpeak': 2.6, 'thal': '7', 'exercise\_induced\_angina': '0', 'maximum\_heart\_rate\_achieved': 157.0\}}

         \noindent $\bullet$ \textbf{Key Proof Tree Points:} 
        \begin{enumerate}
        \item "label(X,'absent') does not hold because the value of \texttt{major\_vessels} is '2' which should equal '0' does not hold."
        \item "label(X,'absent') does not hold because the value of \texttt{chest\_pain} is '4' which should not equal '4' does not hold and the value of \texttt{age} is 60.0 which should be greater than 66.0 does not hold."
    \end{enumerate}

The implementation was done using Python, with libraries such as scikit-learn for ML models, SciPy for statistical tests, and clingo for ASP solving. Pandas and NumPy were used for data manipulation and numerical computations. The FOLD-R++ library was used to induce logical rules from data.
The project was modularized into separate components for data loading, model training, hybrid integration, and evaluation to improve readability.


\section{Results}
\label{sec:results}

This section presents the experimental results of evaluating the hybrid models that integrate black-box ML classifiers with interpretable rules derived from FOLD-R++ using ASP. The performance of the ML models and the hybrid model is compared across multiple medical datasets. Additionally, the role of ASP in improving interpretability and correcting errors is illustrated through case studies.

Table~\ref{tab:combined} summarizes the average accuracy and F1 score of the ML models, FOLD-R++ used in isolation and the hybrid models across all datasets and experiments.

\begin{table}[h!]
\centering
\scriptsize
  \resizebox{0.98\textwidth}{!}{  
\setlength{\tabcolsep}{2pt}
\begin{tabular}{lcccc|| ccc}
\toprule
\textbf{Model} & \textbf{ML Acc} & \textbf{Hybrid Acc} & \textbf{ML F1} & \textbf{Hybrid F1} &   \textbf{t-statistic} & \textbf{p-value} & \textbf{Significant} \\
\midrule
\multicolumn{8}{l}{\textbf{Heart Disease}} \\
\midrule
FOLD-R++ & 74.26 & - & 76.39 & - & - & - & - \\
KNN  & 64.26  & 64.26  & 68.97  & 68.97 & N/A & N/A & N/A \\
MLP  & 80.93  & 80.19  & 81.22  & 80.65 & -0.53 & 0.61 & No \\
RF   & 79.44  & 79.44  & 81.24  & 81.39 & 0.00 & 1.00 & No \\
SVM  & 63.52  & \textbf{71.30}  & 70.96 & \textbf{75.14} & 3.55 & 0.01 & Yes \\
\midrule
\multicolumn{8}{l}{\textbf{Autism Screening}} \\
\midrule
FOLD-R++ & 93.40 & - & 95.49 & - & - & - & - \\
KNN  & 87.38  & 87.38  & 91.52  & 91.52 & N/A & N/A & N/A \\
MLP  & 97.52  & 97.66  & 98.30  & 98.29 & 1.00 & 0.34 & No \\
RF   & 97.38  & 96.10  & 98.20  & 97.33 & -3.04 & 0.01 & Yes \\
SVM  & 72.62  & \textbf{94.04}  & 84.13 & \textbf{96.01} & 32.12 & $1.35 \times 10^{-10}$ & Yes \\
\midrule
\multicolumn{8}{l}{\textbf{Breast Cancer Wisconsin}} \\
\midrule
FOLD-R++ & 94.43 & - & 92.04 & - & - & - & - \\
KNN  & 94.36  & 94.36  & 91.62  & 91.62 & N/A & N/A & N/A \\
MLP  & 93.57  & \textbf{94.43}  & 91.04 & \textbf{92.04} & 3.09 & 0.01 & Yes  \\
RF   & 96.36  & 95.50  & 94.78  & 93.63 & -2.54 & 0.03 & Yes \\
SVM  & 95.21  & 95.36  & 93.11  & 93.28 & 0.80 & 0.44 & No \\
\midrule
\multicolumn{8}{l}{\textbf{Ecoli}} \\
\midrule
FOLD-R++ & 95.15 & - & 94.58 & - & - & - & - \\
KNN  & 65.74  & 65.74  & 57.46  & 57.46 & N/A & N/A & N/A \\
MLP  & 92.50  & 94.71  & 91.14  & 93.93 & 2.75 & 0.02 & Yes \\
RF   & 96.32  & 96.47  & 95.77  & 95.96 & 0.36 & 0.73 & No \\
SVM  & 57.06  & \textbf{87.50}  & 23.37  & \textbf{82.99} & 7.30 & $4.58\times10^{-5}$ & Yes \\
\midrule
\multicolumn{8}{l}{\textbf{Chronic Kidney Disease}} \\
\midrule
FOLD-R++ & 62.00 & - & 76.01 & - & - & - & - \\
KNN  & 89.13  & 89.13  & 90.36  & 90.36 & N/A & N/A & N/A \\
MLP  & 95.25  & 95.50  & 96.05  & 96.28 & 0.69 & 0.51 & No \\
RF   & 100.00 & 99.88  & 100.00 & 99.89 & -1.00 & 0.34 & No \\
SVM  & 91.38  & \textbf{93.13} & 92.60  & \textbf{94.27} & 3.74 & 0.01 & Yes \\
\bottomrule
\end{tabular}}
\caption{Combined results (mean values only) for Accuracy (\%) and F1 Score (\%) across five medical datasets. The constant FOLD-R++ performance for each dataset is indicated in the header. Bold entries highlight the model with the best improvement within each dataset. Statistical significance (p < 0.05) from paired t-tests comparing ML and Hybrid performance is shown on the right (N/A indicates no difference).}
\label{tab:combined}
\end{table}

The results indicate that the hybrid models demonstrated an improvement in accuracy and F1 score over the standalone ML models across several datasets. Paired t-tests, presented in Table~\ref{tab:combined}, were conducted to determine whether these observed improvements were statistically significant. If the ML model and hybrid model have the same accuracy in all experiments, the differences were zero, and the paired t-test could not be performed (denoted as "N/A"). The SVM classifier, in particular, showed significant improvements. When compared to the pure FOLD-R++ approach, which is applied uniformly across each dataset, the hybrid models can significantly enhance performance for classifiers with lower baseline scores.
Conversely, when a baseline model already achieved near-perfect scores, the hybrid approach typically maintained similar performance.

Overall, these findings suggest that the hybrid approach is particularly beneficial for cases where the ML baseline is suboptimal, although the optimal method remains dependent on the dataset.

\subsection{Results by Dataset}

In the following subsections, examples are presented from four representative datasets. For each, we highlight improved predictions (where the ASP rules successfully corrected a misclassification), no-change predictions, and worsened predictions (where the ASP rules overrode a correct classification). For example, a misclassified instance may have triggered a learned ASP rule that encodes clinically relevant conditions, thus overriding a low-confidence ML guess. Conversely, when the conditions or exceptions of a rule were misaligned with the true label, the override led to a worse outcome. Finally, in some high-confidence cases, the model's prediction was already correct and thus remained unchanged.
All examples come from the explanation files generated during the experiments. Detailed analyses for the remaining dataset can be found in Appendix~\ref{app:additional-results}.

\subsubsection{Heart Disease}
The hybrid model improved the accuracy of the SVM classifier from 63.5\% to 71.3\%. In particular, the first rule in Listing~\ref{lst:absent} shows the ASP rule for labeling a case as \emph{absent} (i.e., no heart disease) when \texttt{thal} equals 3 and the maximum heart rate exceeds 71, provided no exceptions apply. 
The second rule in Listing~\ref{lst:absent} defines one such exception (\texttt{ab2}), which triggers if the patient exhibits chest pain type 4 and has a nonzero number of major vessels affected.

\begin{lstlisting}[float=t,
caption={ASP Rules for Heart Disease.},
label={lst:absent}]
label(X, absent) :-    thal(X, 3),    maximum_heart_rate_achieved(X, V_max_hr_1),   
V_max_hr_1 > 71.0,    not ab2(X, True), not ab3(X, True),    not ab4(X, True), not ab5(X, True),    not ab6(X, True), not ab7(X, True),    not ab8(X, True).
ab2(X, True) :-   chest_pain(X, 4),  major_vessels(X, V_major_vessels_1),   V_major_vessels_1 != 0,  not ab1(X, True).
\end{lstlisting}

An illustrative example is \texttt{patient34}, who was originally predicted "absent" by the SVM model. After applying the rule in Listing~\ref{lst:absent}, the hybrid model considered potential exceptions and corrected the prediction to "present". The patient had \texttt{thal} equal to 3 and a maximum heart rate achieved of 150 (greater than 71), but exceptions did not hold due to the absence of chest pain type 4 and no major vessels affected. This alignment with clinical indicators led to a  more accurate diagnosis.

There are cases where the ASP rules did not alter an ML prediction when both the model and the learned rules align on the same label. For example, in one instance the SVM model predicted "present" with high confidence, and the rules confirmed that label without any exceptions triggered.
However, the ASP rules can sometimes degrade performance if they override an already correct ML prediction. This occurs when a learned rule is too general or when exceptions fail to capture the subtle conditions that the ML model detected. In one such instance, the SVM model correctly predicted "present", but the hybrid approach changed it to "absent" because a relevant exception in the rule set was not satisfied.

\subsubsection{Autism Screening}
The SVM classifier showed significant improvement when combined with ASP rules, with accuracy increasing from 72.6\% to 94.0\%. However, performance slightly decreased when hybridized with ASP, likely because of the original high accuracy of the model and the rules occasionally overriding correct predictions.
Listing~\ref{lst:no-autism} presents the rule for labeling an individual as \texttt{NO} (not autistic) if \texttt{a5} is not equal to 1 and exceptions \texttt{ab1}, \texttt{ab2} do not apply. Listing~\ref{lst:no-autism} also outlines the exception rule \texttt{ab1}, which captures strong autistic traits indicated by affirmative answers to certain questions.

\begin{lstlisting}[float=t,
    caption={ASP Rule for Autism Screening.},
    label={lst:no-autism}]
label(X, NO) :-    a5(X, V_a5_0),    V_a5_0 != 1,    not ab1(X, True),    not ab2(X, True).

ab1(X, True) :-   a9(X, 1),    a3(X, 1),    a1(X, 1),    a6(X, 1).
\end{lstlisting}

For instance, in the case of \texttt{patient1}, the ML model predicted "autistic". However, the rule in Listing~\ref{lst:no-autism} applied, and neither \texttt{ab1} nor \texttt{ab2} was valid. Consequently, the hybrid model overruled the ML model and correctly classified \texttt{patient1} as "non-autistic".

Similarly, some instances that the ML model correctly classified as "autistic" were left unchanged by the ASP rules. However, there are also examples in which a correct ML prediction was overridden incorrectly by a default rule or a missing exception. This can happen if the rule induction process relies on partial indicators that conflict with more nuanced patterns captured by the classifier. As a result, the hybrid system incorrectly switched the label to "non-autistic".

\subsubsection{Breast Cancer Wisconsin}
The performance of the MLP classifier improved with the hybrid model, increasing accuracy from 93.6\% to 94.4\%. In contrast, for models that already performed very well such as RF and SVM, the hybrid approach provided minimal or no gains. 
Listing~\ref{lst:malignant-breast} displays a rule labeling a tumor as \texttt{malignant} if cell size uniformity is not 1 and no relevant exceptions arise. Among these exceptions is \texttt{ab1}, shown in Listing~\ref{lst:malignant-breast}, handling cases where, despite non-uniform cell size, other features do not indicate malignancy. For example, \texttt{patient33} was initially misclassified as "benign" by the ML model but was correctly classified as "malignant" by the hybrid model, as \texttt{cell\_size\_uniformity} was equal to 5 (and thus not 1), and additionally, exceptions \texttt{ab1} and \texttt{ab2} did not hold.

\begin{lstlisting}[float=t,
    caption={ASP Rule for Breast Cancer.},
    label={lst:malignant-breast}]
label(X, malignant) :-    cell_size_uniformity(X, V_cell_size_uniformity_0),    V_cell_size_uniformity_0 != 1,    not ab1(X, True),    not ab2(X, True),    not ab3(X, True).
ab1(X, True) :-    bare_nuclei(X, 1),    cell_size_uniformity(X, V_cell_size_uniformity_1),   V_cell_size_uniformity_1 != 10,    marginal_adhesion(X, V_marginal_adhesion_2),    V_marginal_adhesion_2 != 10,    clump_thickness(X, V_clump_thickness_3),    V_clump_thickness_3 != 7,    cell_size_uniformity(X, V_cell_size_uniformity_4),    V_cell_size_uniformity_4 != 6,   bland_chromatin(X, V_bland_chromatin_5),    V_bland_chromatin_5 != 5.
\end{lstlisting}

In this dataset, all correct predictions by the ML model  were confirmed by the hybrid approach. 
Consequently, the hybrid approach provided consistent or improved results.

\subsubsection{Ecoli}
The hybrid model significantly improved SVM accuracy from 57.1\% to 87.5\%. Listing~\ref{lst:cp-ecoli} describes a rule labeling a sample as \texttt{cp} if its sequence name (\texttt{sn}) is not "FECR" and the attribute \texttt{alm1} is less than or equal to 0.38.  Listing~\ref{lst:cp-ecoli} also defines an exception that checks various attributes (\texttt{gvh}, \texttt{mcg}) to invalidate the rule when these features are atypical. 

\begin{lstlisting}[float=t,
    caption={ASP Rule for Ecoli.},
    label={lst:cp-ecoli}]
label(X, cp) :-    sn(X, V_sn_0),    V_sn_0 != FECR,    alm1(X, V_alm1_1),    V_alm1_1 <= 0.38,    not ab1(X, True).

ab1(X, True) :-    sn(X, V_sn_0),    V_sn_0 != PTKB,    gvh(X, V_gvh_1),    V_gvh_1 > 0.55,    mcg(X, V_mcg_2),    V_mcg_2 > 0.41.
\end{lstlisting}

For example, \texttt{patient48} was classified as "negative" by the ML model. As the first rule in Listing~\ref{lst:cp-ecoli} applied, hence, the hybrid model corrected the classification to "positive".

In several examples, if the ML model and the ASP converged on the same conclusion, the final prediction remained unchanged. On the other hand, the hybrid system can misclassify instances when a rule is activated at the wrong threshold or lacks the proper exception. In one instance, for example, the SVM model accurately predicted "negative", but the hybrid system changed the prediction to "positive".

\subsection{Explainability Analysis}
In addition to performance improvements, our hybrid approach enhances explainability through interpretable ASP rules, which we examine in this section.

When ML models make correct predictions with high confidence, the hybrid model confirms these predictions while adding transparent explanations. For instance, in the Heart Disease dataset (\texttt{patient4}), the SVM model correctly predicted "present" with 0.77 confidence. The hybrid model confirmed this, while explaining that although "absent" rule was partially activated, the critical exceptions were not met.

A key consideration is explainability when the ML classifier and FOLD-R++ disagree. In the Breast Cancer dataset ((\texttt{patient22}), the MLP model correctly predicted "malignant" with high confidence (nearly 1.0), whereas the FOLD-R++ incorrectly predicted "benign". The hybrid model maintained the correct ML prediction based on confidence thresholding. The ASP explanation for this instance indicates that although the rule for a malignant classification was partially activated, several critical conditions were not met. Consequently, the rule did not fully trigger a malignant classification. 
This case highlights a limitation in our approach: while the system correctly maintained the ML prediction due to confidence thresholding, the available ASP explanation would have supported an incorrect prediction. Thus, there is a trade-off between interpretability and accuracy.

Conversely, in the Autism Screening dataset (\texttt{patient2}), the SVM model incorrectly predicted "autistic" with 0.01 confidence, while FOLD-R++ correctly predicted "non-autistic". Consequently, the hybrid model overrode the uncertain ML output. The ASP explanation for this instance shows that the rule for a "non-autistic" diagnosis was activated and other condition checks were met, while no exception conditions were triggered. Here, the ASP explanation not only corrected the ML prediction but also provided a clear rationale that could be validated by domain experts.

These examples demonstrate that while our approach generally enhances explainability, explanation reliability depends on rule quality, confidence thresholds, and alignment with domain knowledge. When ML classifiers have high accuracy, the added value of ASP explanations may be limited from a performance perspective, but they still increase interpretability. For lower-accuracy ML models, ASP rules can simultaneously improve performance and provide reliable explanations.


\section{Conclusion}
\label{sec:conclusion}

This study investigated the integration of interpretable ASP rules derived from the FOLD-R++ algorithm with black-box ML models to improve predictive performance and interpretability in medical classification tasks. A central advantage demonstrated by this method is its ability to integrate ASP-derived rules without modifying the internal mechanisms of the classifiers. As shown in the experiments, this offers a practical way to combine robust ML predictions with explainable rules that can correct low-confidence or erroneous outputs.
The study addressed two research questions.

First, regarding whether integrating interpretable ASP rules improves the predictive performance of black-box ML models across various medical datasets, the results demonstrate significant improvements. The hybrid model notably improved the accuracy and F1 scores of several ML classifiers, particularly the SVM, across multiple datasets. For example, in the Autism Screening Dataset, the hybrid model increased the accuracy of the SVM classifier from 72.6\% to 94.0\% (p < $ 10^{-10}$). Similarly, in the Ecoli dataset, the accuracy improved from 57.1\% to 87.5\% (p < $0.0001$). This suggests the hybrid model was able to address limitations in certain ML models when dealing with complex or noisy data. In general, the largest improvement where seen on sub-optimal ML classifiers.

The results demonstrate that while the hybrid approach significantly enhances performance for classifiers with lower baseline scores, there are also datasets where standalone FOLD-R++ outperforms both the baseline ML and the hybrid methods. This reflects our core research objective of enhancing ML classifiers, rather than asserting that the hybrid approach exceeds all alternatives.
Notably, even when ML models perform better than standalone FOLD-R++, as in the CDK dataset, the hybrid approach can still offer incremental improvements while preserving interpretability.

Second, concerning how the hybrid model improves the interpretability of predictions, the integration of ASP rules provides clear, human-readable explanations for each prediction. The logical rules derived from the FOLD-R++ algorithm align with domain knowledge, and make it easier to understand and trust the predictions. For instance, in the Chronic Kidney Disease dataset, the rule for labeling "ckd" reflects the medical understanding that elevated serum creatinine levels (\texttt{sc}) are indicative of kidney dysfunction. The rules improved the accuracy of predictions and provided an  explanation aligned with medical knowledge, which improves transparency and trust.

Although the current implementation does generate an ASP-based proof tree for every instance, the primary advantage occurs where the hybrid system actually overrides a low-confidence or misclassified ML prediction. In those cases, the symbolic justification is crucial for conveying why the override applies. In high-confidence cases that the ML model already classifies correctly, the hybrid model typically does not rely on the proof tree to alter the outcome, though an explanation is still recorded.
By contrast, if the prediction is already high-confidence and correct, the system does not rely on the proof tree to alter the outcome.  If every prediction requires a detailed rationale, one could incorporate additional post-hoc explainers  (e.g.\ SHAP-FOLD \cite{shakerin2019induction}).


However, while the hybrid model shows promising results, several limitations should be considered. As datasets become larger and more complex, the number of induced ASP rules can increase substantially, which could affect interpretability. 
Additionally, heavy use of exceptions for capturing specific cases may cause the rules to overfit the training data.
The variations in induced rules across different experiments suggest sensitivity of the FOLD-R++ algorithm to data splitting and randomness. 
Furthermore, the hybrid model did not consistently improve performance across all ML models and datasets, as the benefits for KNN and RF on certain datasets were minimal, indicating the effectiveness of the hybrid approach depends on the characteristics of both the dataset and base ML model.

Finally, running an ASP solver like Clingo introduces additional computational overhead. For larger datasets, this may require further optimizations. 

Future work should extend this research across several dimensions. A primary focus will be on enhancing scalability, particularly by addressing the computational overhead of ASP solvers like Clingo on larger datasets and exploring alternative symbolic engines. Broader empirical evaluations on more diverse, larger datasets and applications to new domains will also be pursued, incorporating a wider variety of ML architectures to benchmark against state-of-the-art models. Systematic optimization of the confidence threshold and further investigation into the effects of data transformation will be pursued, including assessing FOLD-R++'s capabilities with less transformed or original data. To address the variability observed in induced rules and improve their robustness, dynamic rule refinement techniques, such as ILP, will be explored. Further research includes human-in-the-loop evaluations with domain experts to validate interpretability and clinical utility, and the investigation of more advanced hybrid architectures.

%% file: appendix.tex
\section{Additional Dataset Results}
\label{app:additional-results}

This appendix provides detailed results and case study discussions for the Chronic Kidney Disease dataset, supplementing the main analysis in Section~\ref{sec:results}.

\subsection{Chronic Kidney Disease (CKD)}
For the SVM classifier, the hybrid model improved accuracy from 91.4\% to 93.1\%. Listing~\ref{lst:ckd-kidney} shows the rule which flags elevated serum creatinine levels as indicative of CKD. 
For example, \texttt{patient70} was incorrectly predicted as "not ckd" by the SVM model. The hybrid model corrected this to "ckd" as the  \texttt{sc} value (1.3) exceeds the threshold.

\begin{lstlisting}[float=h,
    caption={ASP Rule for Labeling \texttt{ckd} (Chronic Kidney Disease).},
    label={lst:ckd-kidney}]
label(X, ckd) :-    sc(X, V_sc_0),   V_sc_0 > 1.2.
\end{lstlisting}

In many instances, the ML model and the ASP rules independently converge on the same prediction. In such cases, the hybrid model simply confirmed the ML prediction. As none of the instances were overridden incorrectly by the rules, the hybrid model never worsened the outcomes. 
Interestingly, the RF classifier achieved perfect accuracy (100\%) on this dataset, while the hybrid approach averaged 99.9\%. This slight decrease occurs when ASP rules override correct RF predictions made with confidence below the threshold, highlighting a potential trade-off for near-optimal base models.